\documentclass[10pt,twocolumn,letterpaper]{article}

\usepackage{cvpr}
\usepackage{times}
\usepackage{epsfig}
\usepackage{graphicx}
\usepackage{amsmath}
\usepackage{amssymb}
\usepackage{enumitem}

\usepackage[breaklinks=true,bookmarks=false]{hyperref}

\cvprfinalcopy 

\def\cvprPaperID{****} 

\begin{document}

\title{Products-10K: A Large-scale Product Recognition Dataset}

\author{Yalong Bai, Yuxiang Chen, Wei Yu, Linfang Wang, and Wei Zhang\\
JD AI Research\\
{\tt\small \{baiyalong, chenyuxiang9, yuwei10, wanglinfang, zhangwei96\}@jd.com}
}

\maketitle

\begin{abstract}
With the rapid development of electronic commerce, the way of shopping has experienced a revolutionary evolution. To fully meet customers' massive and diverse online shopping needs with quick response, the retailing AI system needs to automatically recognize products from images and videos at the stock-keeping unit (SKU) level with high accuracy. However, product recognition is still a challenging task, since many of SKU-level products are fine-grained and visually similar by a rough glimpse. Although there are already some products benchmarks available, these datasets are either too small (limited number of products) or noisy-labeled (lack of human labeling). In this paper, we construct a human-labeled product image dataset named ``Products-10K'', which contains 10,000 fine-grained SKU-level products frequently bought by online customers in JD.com. Based on our new database, we also introduced several useful tips and tricks for fine-grained product recognition. The products-10K dataset is available via \url{https://products-10k.github.io/}.

\end{abstract}

\section{Introduction}
Products, the most common objects in our daily life, are deeply related to everyone through commercial activities. Recognizing daily products by snapping a photo is one of the most fundamental problems in pattern recognition, for both academia and industry. Especially in the age of online shopping, it is desirable and convenient for an AI system to recognize products accurately and efficiently, from a massive amount of SKUs (stock keeping unit). Until now, there are already lots of SKU-level product recognition related application scenes, such as online snap-shopping, on-device supermarket product recognition, Amazon Go-style cashier-less stores, etc.

\begin{figure}[t]
\centering
\includegraphics[width=\linewidth,page=1]{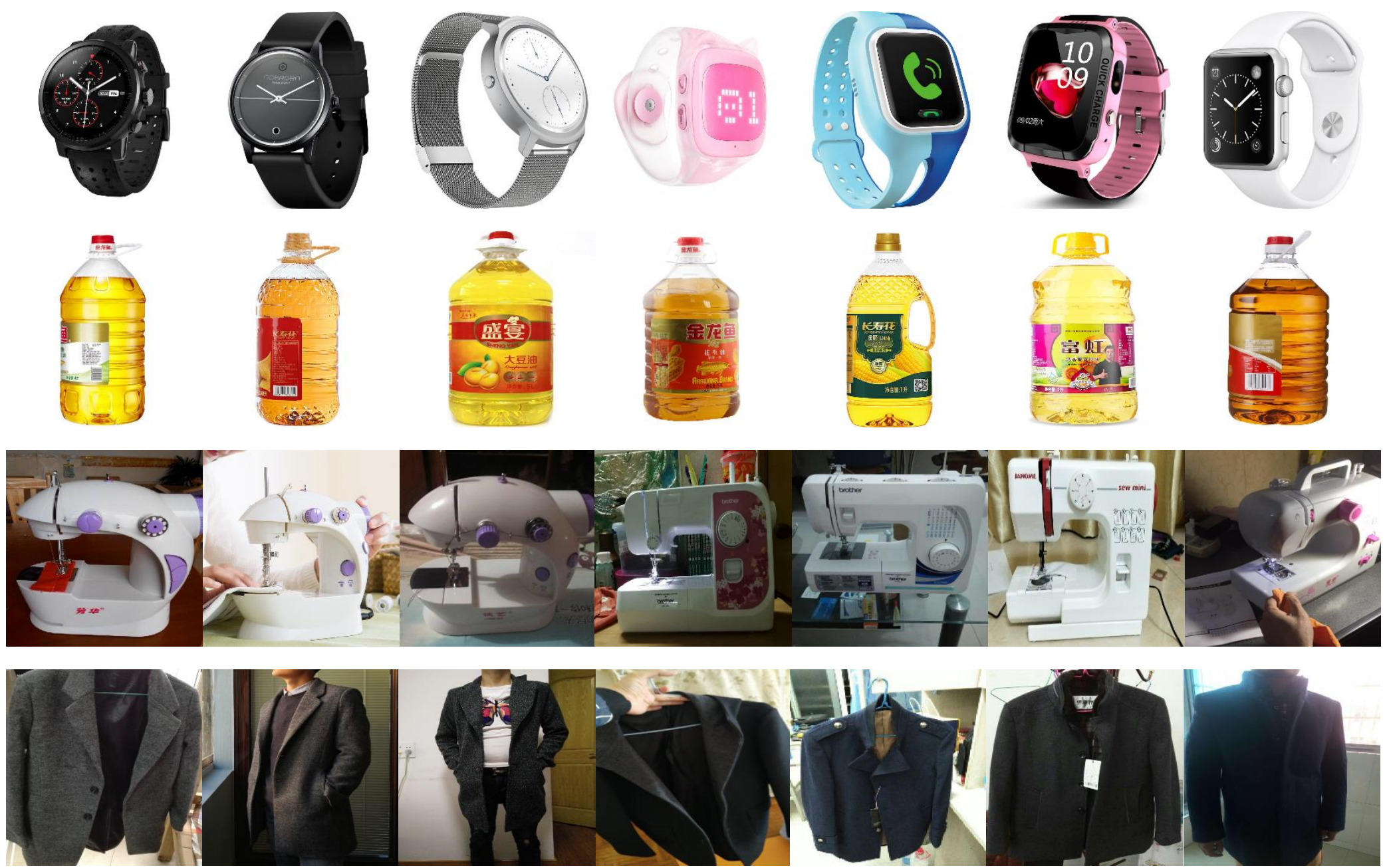}
   \caption{Examples of SKUs in Product-10K dataset. Images in each row is visually similar to each others, but they belong to different SKUs.}
\label{fine-grained}
\end{figure}

The SKU-level product recognition is different from generic object recognition, since most of the SKUs are visually similar by a rough glimpse, as shown in Figure~\ref{fine-grained}. It can be regarded as a fine-grained image recognition task, and these visually similar SKUs can be correctly recognized by details in discriminative local regions. As we knew, fine-grained image recognition~\cite{chen2019destruction,zhou2020look} has made rapid progress owing to the development of deep-learning-based technologies. Image datasets with SKU labels serve as the backbone for such data-driven deep model training~\cite{hu2020diffnet,tonioni2018deep}.

\begin{figure*}[t]
\centering
\includegraphics[width=1\linewidth,page=1]{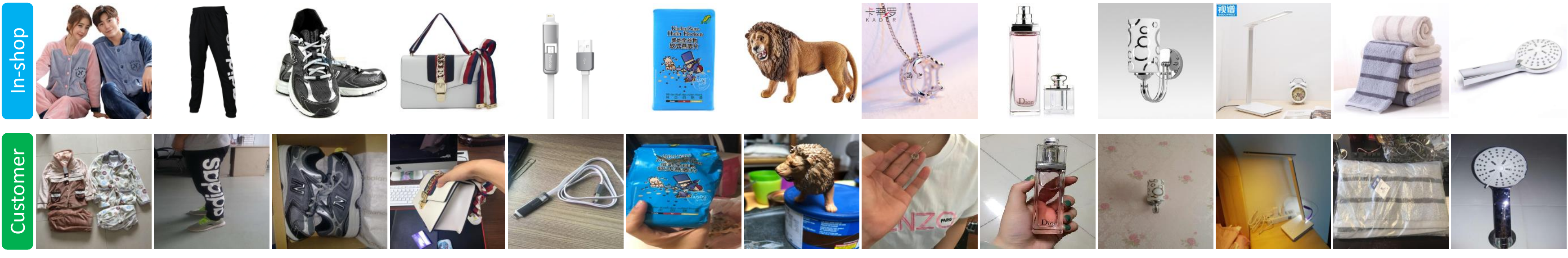}
   \caption{Images in Product-10K are collected from in-shop photos (the first row) and customer images (the second row). The SKU label of images in each column is same.}
\label{two_groups_image}
\end{figure*}

In general, the product images can be grouped into two sets: \textit{in-shop} photos and \textit{customer} images, as shown in Figure~\ref{two_groups_image}. In-shop photos are usually object-centric with standard shooting angles and clear background. These kinds of images are friendly to guide a fine-grained recognition model to detect the discriminative regions among different SKUs. While customer images have complex backgrounds with irregular quality. Due to the various camera angles and ambient light, recognizing produce from customer images are thought to be more difficult. But training from these kinds of images benefit to improve the generalization ability of the product recognition models. Therefore, both of these two kinds of images are essential for learning a high-performance product classifier. However, nearly 44.5\% of customer images are noise data, which is irrelevant to their labeled SKUs. The deep convolutional neural networks trained on a high percentage of noise tends to a significant performance drop~\cite{sukhbaatar2014training}.

In this paper, we constructed a human-labeled product recognition dataset named ``Products-10K'', which is so far the largest dataset containing nearly 10,000 SKUs frequently bought by online customers in JD.com\footnote{The largest online retailer in China}. We aim to bring the fine-grained SKU level recognition of daily products into a computer vision community. Both of the in-shop photos and customer images are collected in this Products-10K, the noise rate of this dataset is lower than 0.5\%.

Moreover, we examine a collection of basic tricks and tips for SKU-level product recognition, including high-resolution training, unbiased learning from bias dataset, metric-guided loss function design that improves model accuracy without introducing additional computation cost. All of these tips and tricks have been demonstrated with great performance improvements on various SKU-level product recognition tasks. These key findings have also been applied in the winner solution of previous ``AliProducts Challenge: Large-scale Product Recognition at CVPR 2020'' and ``iMaterialist Challenge on Product Recognition at CVPR 2019''.

\section{Related Work}
We list existing known competitions or datasets related to fine-grained production recognition below:
\begin{enumerate}[itemsep=0pt,parsep=0pt,topsep = 0pt]
\item iMaterialist Competition\footnote{https://www.kaggle.com/c/imaterialist-challenge-FGVC2017}, FGVC4, CVPR2017.
\item iMaterialist Competition\footnote{https://www.kaggle.com/c/imaterialist-challenge-furniture-2018}, FGVC5, CVPR2018.
\item iMaterialist Challenge on Product Recognition\footnote{https://www.kaggle.com/c/imaterialist-product-2019}, FGVC6, CVPR2019.
\item AI Meets Beauty\footnote{https://challenge2019.perfectcorp.com/}. ACM Multimedia 2018 and 2019.
\item JD AI + Fashion Challenge\footnote{https://fashion-challenge.github.io/}, ChinaMM 2018.
\item AliProducts Challenge: Large-scale Product Recognition\footnote{https://retailvisionworkshop.github.io/}, RetailVisual Workshop, CVPR2020.
\end{enumerate}

However, all of the above competitions have several apparent limitations. The iMaterialist Competition at FGVC4 is designed for recognizing 381 attribute labels for productions rather than SKU-level production labels. The iMaterialist Competition at FGVC5 only contains the categories of a sub-domain of productions (Furniture). Compared to the iMaterialist Challenge on Product Recognition at FGVC6, which contains 2019 SKUs, we have a much larger set of SKUs (10k versus 2k). Moreover, all images in our dataset are fully annotated by human experts. While the training set of iMaterialist Challenge on Product Recognition at FGVC6 and AliProducts Challenge at RetailVisual Workshop is not manually-labeled and contains a large proportion of noisy images. Moreover, compared to AI Meets Beauty Challenge and JD AI+Fashion Challenge that focus only on Fashion products, our competition covers much larger categories, including Fashion, 3C, food, healthcare, household commodities, etc.. 

In general, compared with previous product recognition dataset, Products-10K has three obvious advantages: 1) large-scale dataset covering 10k fine-grained SKU-level products active on e-commerce website; 2) covers a full spectrum of products, rather than only a small domain (e.g., Fashion, Furniture); 3) fully annotated with human experts, without noisy labels. 

\section{Products-10K Dataset}
\begin{figure*}[!ht]
\centering
\includegraphics[width=\linewidth,page=1]{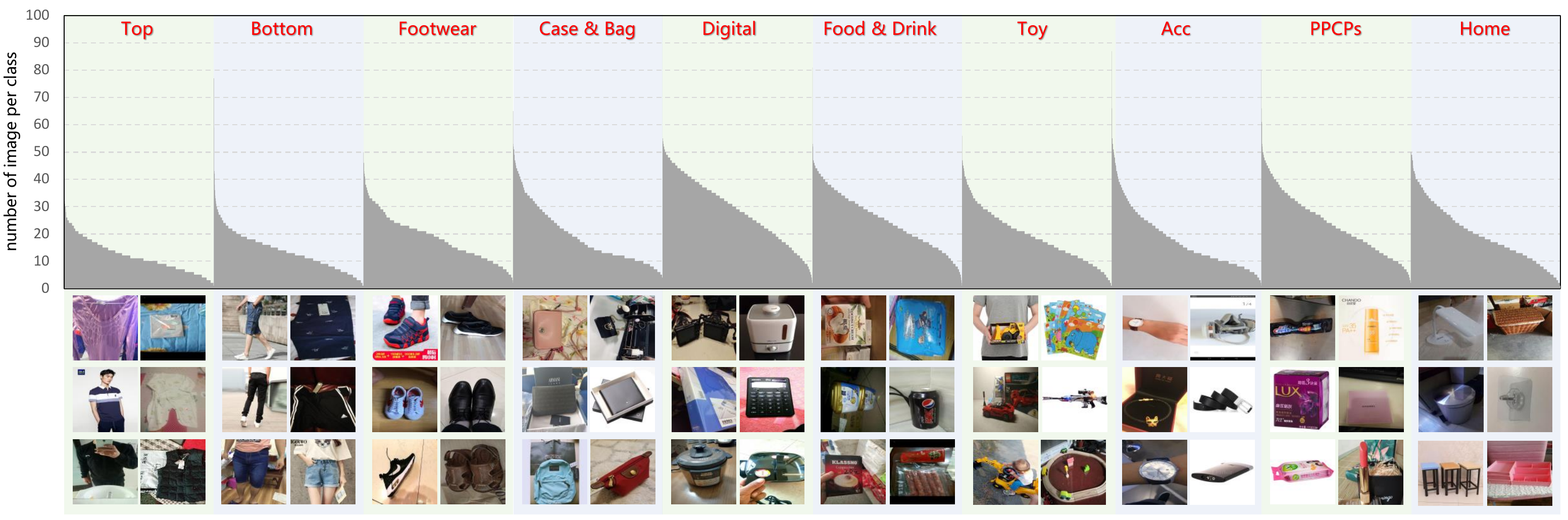}
\caption{Examples of images in Products-10K dataset. Both of the clean in-shop photos and realistic customer images are collected.}
\label{distribution}
\end{figure*}

All of the images in our Products-10K dataset are collected from the online shopping website JD.com. Ten thousand frequently bought SKUs are contained in the dataset, covering a full spectrum of categories including Fashion, 3C, food, healthcare, household commodities, etc.. Moreover, large-scale product labels are organized as a graph to indicate the complex hierarchy and inter-dependency among products. There are nearly 150,000 images in total. Respect to the practical application scenes, the distribution of image amount is unbalanced, as shown in Figure~\ref{distribution}. 

\textit{All images are manually checked/labeled by the production expert team of JD.com.} Each image is checked whether its label is wrong or not by at least three human experts. There are nearly 44.5\% of noise customer images are filtered out by human experts. The noise rate of the whole dataset is low than 0.5\%. Both of the in-shop photos and customer images are collected for each SKU in Products-10K. Customer images are more difficult for recognizing owing to the complex background, color distortion, various lighting, and angle of view in the images. It makes the product recognition on Products-10K to be a more challenging task due to the domain differences of images in the datasets. The products-10K dataset has already been available for non-commercial research and educational purposes. 

\section{Proposed Methods}
\paragraph{Training on High-resolution Image} SKUs-level product recognition is a fine-grained visual categorization task, discriminative feature representation learning play a very important role in distinguishing visually similar categories. Standard object recognition deep models usually resize all input image to 256$\times$256 and random crop 224$\times$224 for training and then resize images to 256$\times$256 and center crop 224$\times$224 for test. Such low-resolution input is usually sufficient for distinguishing structural or semantic variances among images, but it has limited capability for describing the detail differences among visually similar SKUs. 
A reasonable solution is to directly train the product recognition model on high-resolution images.
Here we resize input image to 512$\times$512 and random crop 448$\times$448 for model finetuning. In this way, detail discriminative regions in images can be easily captured by the recognition model.

\paragraph{Balanced Subset Finetune} Training with an unbalanced dataset leads the models to achieve basic performance in classification, and results in performance drop in a balanced validation set due to the bias in the model. Although there are already some optimization methods have proposed for handling class imbalanced data, such as class-balance loss function~\cite{aurelio2019learning,cui2019class}, or weighted resampling~\cite{Li_2019_CVPR}. However, a more simple and effective method is to finetune the model trained on full set by using a balanced subset of the training dataset. In AliProduct Challenge, such a solution can lead to a dramatic improvement in various backbone networks trained on the fullset. For the products-10K dataset, we randomly sample at most 5 images (using all images if the number of images in the corresponding category is less than 5) for each SKUs to construct a balanced subset for model finetuning.

\begin{figure}[!ht]
\centering
\includegraphics[width=\linewidth,page=1]{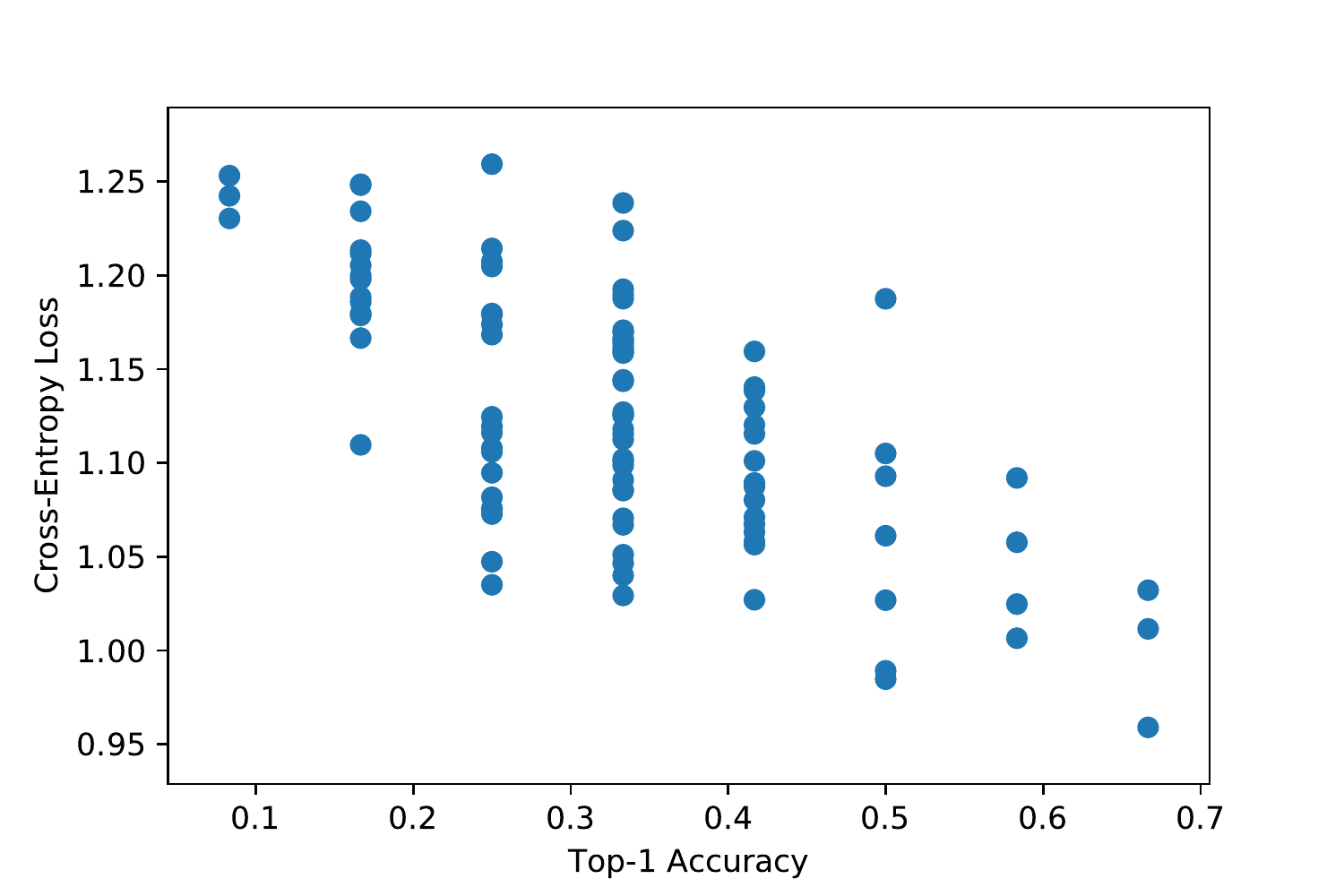}
\caption{The distributions of cross-entropy loss and top-1 accuracy score. Given 12 ground-truth samples from a three-label classification dataset, we randomly sampled 1,000 different possible predictions for each sample.}
\label{fig:acc1_loss}
\end{figure}

\paragraph{Metric Loss Function} As we known, most of the recognition models are optimized by the cross-entropy loss functions, while the practical evaluation metric is usually the top-1 accuracy on the test set. However, if we try to calculate cross-entropy loss for all the possibilities and compare them with top-1 accuracy, we can find many examples, when loss differs while the top-1 accuracy metric stays same, or some other examples whose loss are same but their top-1 accuracy differs, as shown in Figure~\ref{fig:acc1_loss}. The misalignment between the objective function and the final metric method can result in suboptimal convergence. 

Here we design an accuracy loss $L_{acc}$ to optimize accuracy in each batch. The loss can be defined as following:
\begin{equation}
    L_{acc} = 1 - \frac{sum(\hat{y} * y)}{N}
\end{equation}
where $N$ is the size of min-batch, $y$ and $\hat{y}$ indicate the prediction and ground truth in one-hot representation, and $y$ is  the probability outputs which has been normalized by softmax function. We use this accuracy loss finetune the model in 2-3 epoch in the balanced train subset.

\section{Experiments}
We evaluate the performance of our proposed tricks on the validation set of Products-10K.

EfficientNet-B3~\cite{tan2019efficientnet} pre-trained on ImageNet classification task is selected as our backbone due to its generality and effectiveness in recognition tasks, and then we applied our proposed tricks above-mentioned one-by-one on EfficientNet-B3. The optimization is performed using ADAM~\cite{kingma2014adam} with weight decay $1e-4$ and a minibatch size of 64. We use learning rates of $3e-4$ and $3e-5$ for the model trained from scratch and model finetuning respectively. The learning rates are decayed by a factor of 0.1 after 20,40,60 epochs. We trained model based on 448$\times$448 resolution image which is randomly cropped from the resized 512$\times$512 images. Standard data augmentation policies~\cite{krizhevsky2012imagenet} that applied in ImageNet recognition task are directly used in our experiments.

The final experimental results are shown in Table~\ref{tab:res}. It can be found that all of the tricks introduced in this paper can progressively outperform the standard recognition baseline with a large margin. Especially, training from high-resolution gains 4.9\% improvement on top-1 accuracy over the baseline model trained from 224$\times$224 images. With metric loss function finetuned from a balanced subset, the performance can be further improved 1.8\%. By using these tricks, the recognition ability of the model is enhanced but without introducing additional inference cost.

\begin{table}[]
    \begin{center}
    \begin{tabular}{|l|c|}
    \hline
        Method & Accuracy (\%) \\\hline\hline
        Efficient-B3 (224$\times$224) & 60.04 \\\hline
        Efficient-B3 (448$\times$448)   & 62.98\\
        + balanced subset finetune & 63.36\\
        + metric loss function & 64.12\\
        \hline
    \end{tabular}
    \end{center}
    \caption{Top-1 accuracy of a single model on the Products-10K validation set by using a single central crop for testing.}
    \label{tab:res}
\end{table}

\section{Conclusion}
In this paper, we constructed a large-scale product recognition dataset, which is human-labeled and covering nearly 10,000 frequently bought SKUs. Moreover, several useful model finetuning tricks are introduced. We also host a challenge based on this dataset, more detail can be found in \url{https://www.kaggle.com/c/products-10k}.

{\small
\bibliographystyle{ieee_fullname}
\bibliography{egbib}
}

\end{document}